\documentclass[conference]{IEEEtran}
\IEEEoverridecommandlockouts

\usepackage{cite}
\usepackage{amsmath,amssymb,amsfonts}
\usepackage{algorithmic}
\usepackage{graphicx}
\usepackage{textcomp}
\usepackage{xcolor}
\usepackage{hyperref} 

\def\BibTeX{{\rm B\kern-.05em{\sc i\kern-.025em b}\kern-.08em
    T\kern-.1667em\lower.7ex\hbox{E}\kern-.125emX}}
\begin{document}

\title{SpasticMyoElbow: Physical Human-Robot Interaction Simulation Framework for Modelling Elbow Spasticity}

\author{Hao Yu$^{1,2,\dag}$, Zebin Huang$^{1,2}$, Yutong Li$^{3}$, Xinliang Guo$^{4}$, Vincent Crocher$^{4}$,\\ 
Ignacio Carlucho$^{1}$, Mustafa Suphi Erden$^{1}$ 
\thanks{*This work has been submitted to the IEEE for possible publication. Copyright may be transferred without notice, after which this version may no longer be accessible.}
\thanks{**This study was funded by the EPSRC Centre for Doctoral Training in Robotics and Autonomous Systems under the grant reference EP/S023208/1.}
\thanks{$^{1}$School of Engineering and Physical Sciences, Heriot-Watt University, Edinburgh, UK.}%
\thanks{$^{2}$School of Informatics, the University of Edinburgh, Edinburgh, UK.}
\thanks{$^{3}$School of Engineering, the Hong Kong University of Science and Technology, Hong Kong, China.}
\thanks{$^{4}$Department of Mechanical Engineering, the University of Melbourne, Melbourne, Australia.}
\thanks{$^{\dag}$ Hao Yu is the corresponding author, e-mail: {\tt\small s2273711@ed.ac.uk}}}

\maketitle

\begin{abstract}
Robotic devices hold great potential for efficient and reliable assessment of neuromotor abnormalities in post-stroke patients. However, spasticity caused by stroke is still assessed manually in clinical settings. The limited and variable nature of data collected from patients has long posed a major barrier to quantitatively modelling spasticity with robotic measurements and fully validating robotic assessment techniques. This paper presents a simulation framework developed to support the design and validation of elbow spasticity models and mitigate data problems. The framework consists of a simulation environment of robot-assisted spasticity assessment, two motion controllers for the robot and human models, and a stretch reflex controller. Our framework allows simulation based on synthetic data without experimental data from human subjects. Using this framework, we replicated the constant-velocity stretch experiment typically used in robot-assisted spasticity assessment and evaluated four types of spasticity models. Our results show that a spasticity reflex model incorporating feedback on both muscle fibre velocity and length more accurately captures joint resistance characteristics during passive elbow stretching in spastic patients than a force-dependent model. When integrated with an appropriate spasticity model, this simulation framework has the potential to generate extensive datasets of virtual patients for future research on spasticity assessment.
\end{abstract}


\section{Introduction}
Statistics from the Global Burden of Disease Study show that there were approximately 12.2 million new stroke cases and 101 million prevalent cases worldwide in 2019 \cite{feigin2021global}. Spasticity is a common complication following stroke, referred to as a type of abnormal neuromuscular activity, with an incidence of 17\%-38\% in post-stroke patients \cite{SpaDiaAndManHand2016Why}. Given the large number of stroke cases, the prevalence of spasticity is significant in society. The cost of treating stroke might be fourfold due to the influence of spasticity, making the assessment and therapy of spasticity an important research direction in the post-stroke rehabilitation \cite{lundstromFourfoldIncreaseDirect2010}.

Pandyan et al. defined spasticity as a symptom of disordered sensorimotor control caused by an upper motor neuron (UMN) lesion, featuring intermittent or sustained involuntary activation of muscles \cite{Pandyan2018definition}. They stratified the all-encompassing definition into multiple positive features of the UMN syndrome for practical study or measurement, including abnormal activation of muscles to an externally imposed stretch, spasms, clonus, and spastic dystonia \cite{Pandyan2018definition}. In this paper, the term spasticity specifically refers to abnormal muscle reflexes to external stretches in elbows.

In most clinical settings, clinicians use manual scales, such as the Modified Ashworth Scale, to assess spasticity, but these scales struggle with subjective judgement, ambiguous terminology, measurement errors, and inconsistencies in assessment results \cite{MTSandMAS2014}.  Consequently, a range of robotic devices and instrumented assessment approaches have been developed to evaluate spasticity using biomechanical and neurophysiological measures \cite{RobotSpasticity2022, myReview2024}. Beyond merely extracting mathematical metrics from instrumented data, many quantitative models of spasticity are proposed to describe spasticity conditions with model parameters \cite{chaQuantitativeModelingSpasticity2020}. For instance, impedance models can identify the viscoelasticity of spastic elbow joints \cite{parkHapticRecreationElbow2011a} and threshold models represent the impaired ability to regulate the tonic stretch reflex thresholds \cite{calotaSpasticityMeasurementBased2008}.

Aiming to provide a more explainable approach to spasticity characterisation, some studies proposed neuromuscular spasticity models grounded in pathophysiology and realised numerical simulation. He et al. \cite{heDynamicNeuromuscularModel1997} developed the first neuromusculoskeletal model to simulate the pendulum test for knee spasticity, incorporating dynamic knee models driven by five musculotendon actuators, motoneuron activation, and the stretch reflex pathway. Feng et al. \cite{fengNeuromuscularModelStretch1998} adapted He et al.'s model to simulate the pendulum test for elbow spasticity, and validated their simulation through experimental data of joint trajectories and muscle reflex activations from spastic patients. However, the models based on pendulum tests are unsuitable for patients with severe spasticity \cite{schmitReflexTorqueResponse1999}, as high muscle tone often causes their elbows to remain flexed. As a result, Koo et al \cite{kooNeuromusculoskeletalModelSimulate2006}. developed a neuromusculoskeletal model to simulate the stretch reflex response triggered by constant-velocity stretches in a spastic elbow. They innovatively introduced the firing characteristics of muscle spindles into the model and represented the input-output relationship of the $\alpha$-motoneuron pool using a Gaussian cumulative distribution function. 


With the advent of OpenSim \cite{delpOpenSimOpensourceSoftware2007}, which offers well-developed and validated musculoskeletal models, muscle dynamics, and simulation algorithms of human motion, researchers have developed feedback-dependent spasticity models in OpenSim to simulate the effects of spasticity during both passive and active leg movements. Krogt et al. \cite{vanderkrogt2016Neuromusculoskeletal} proposed a velocity-dependent reflex model to simulate joint trajectories of a spastic knee during passive stretches, while, in subsequent research, Falisse et al. \cite{falisseSpasticityModelBased2018} demonstrated that a force-dependent reflex model outperformed a velocity-based model, as it better aligned estimated muscle activity with experimental data of passive stretches and gait in children with spasticity. More recently, Veerkamp et al. \cite{veerkamp2023predicting} contradicted these findings, reporting that a velocity-based model outperformed a force-based model in simulating gait patterns of spastic patients. A detailed comparison of their modelling approaches and open-source code revealed that, while these studies all employed OpenSim, they used distinct optimisation algorithms in different simulation frameworks \cite{vanderkrogt2016Neuromusculoskeletal, falisseSpasticityModelBased2018}. Krogt et al. \cite{vanderkrogt2016Neuromusculoskeletal} obtained the optimal model parameters through matching predicted muscle excitability with experimental data, whereas Falisse and Veerkamp et al. \cite{falisseSpasticityModelBased2018, veerkamp2023predicting} realised the optimisation by fitting joint trajectory data. It is likely that their models achieved only a 'local optimum' within their respective datasets, optimisation targets, and simulation frameworks. In addition, these OpenSim-based spasticity simulation methods have so far been implemented and tested only on leg models, and they were computationally expensive, especially for testing combinations of spasticity model types \cite{veerkamp2023predicting}.  

\begin{figure}[t]
    \centering
    \includegraphics[width=0.45\textwidth]{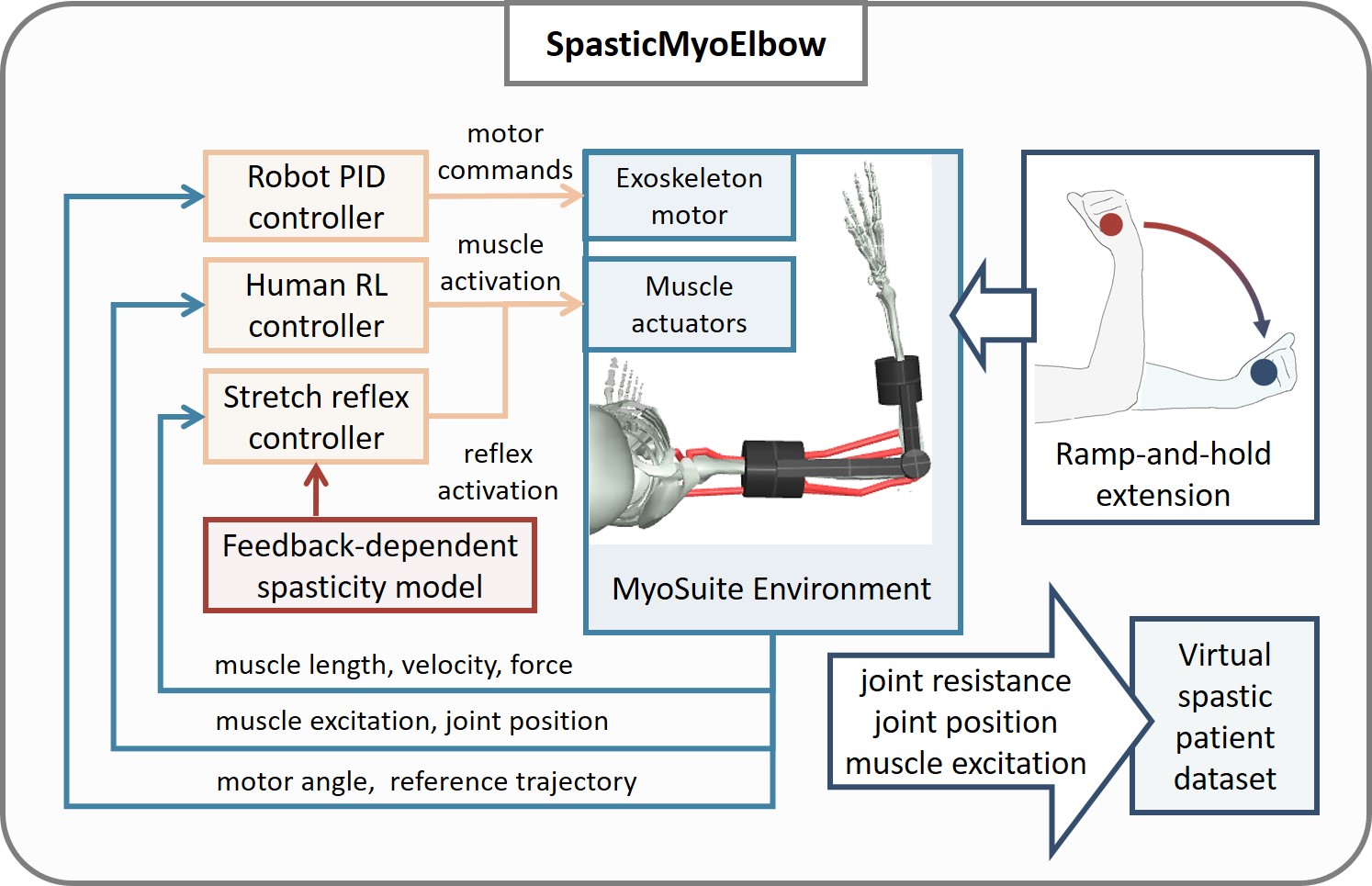}
    \caption{Framework diagram of SpasticMyoElbow. The physical interaction between the robot and the musculoskeletal model during a ramp-and-hold extension can be simulated through the parallel control of the robot and human controllers within a visual simulation environment. The stretch reflex controller allows the elbow model to exhibit spasticity symptoms, simulating a virtual patient in the interaction. This setup provides a validation dataset for evaluating spasticity models.}
    \label{fig:tool}
    \vspace{-5mm}
\end{figure}

In this paper, we present \href{https://github.com/myo-manipulation/SpasticMyoElbow.git}{SpasticMyoElbow}\footnote{The repository can be accessed at: \url{https://github.com/myo-manipulation/SpasticMyoElbow.git}.}, a novel simulation framework for spasticity modelling that enables forward simulation of physical human-robot interaction in robot-assisted spasticity assessment, providing a transparent, practical, and physiologically grounded platform for validating and comparing different elbow spasticity models. This framework was implemented using MyoSuite \cite{MyoSuite2022}, the latest simulator for musculoskeletal system motion. We integrated a robotic controller and a stretch reflex controller into an exoskeleton-elbow environment provided by MyoSuite to simulate physical interactions occurring during robot-assisted elbow spasticity assessments (see Fig.\ref{fig:tool}). With this simulation tool, we reproduced the constant-velocity stretch experiment commonly used for spasticity assessment on an elbow musculoskeletal model, evaluating four types of spasticity models. Our findings show that a feedback-dependent spasticity model combining muscle length and velocity is the most suitable for simulating joint resistance responses of elbow spasticity. The main innovations of this study include: 
\begin{itemize}
    \item Developing an open and flexible simulation framework for testing and comparing elbow spasticity models;
    \item Determining the best model to describe the passive stretch resistance of spastic elbows;
    \item Exploring the application of physical interaction simulation between virtual patients and rehabilitation robots.
\end{itemize} 


\section{Methods}
This work expanded on our prior study \cite{yu2024WCCI}, in which we built a simulation tool for physical interaction between an elbow and an exoskeleton in MyoSuite to study the human similarity of a reinforcement learning (RL) agent in controlling an elbow musculoskeletal model. We newly added a stretch reflex controller unit into the simulation tool to enable the simulation of spasticity models on the virtual elbow. Afterwards, we simulated the robot-assisted constant-velocity stretch experiment for four types of spasticity models to explore the optimal modelling method of elbow spasticity.

\subsection{Simulation framework}
The foundation of the simulation framework is the simulation environment of a typical clinical setup for upper-extremity rehabilitation \cite{schmitReflexTorqueResponse1999, kooNeuromusculoskeletalModelSimulate2006, guo2022practical}, in which a seated person’s elbow interacts with an exoskeleton robot, as shown in Fig.\ref{fig:tool}. This simulation environment was adapted from the elbow-exoskeleton model provided by MyoSuite \cite{MyoSim2022}, which contains a human elbow skeleton with six muscle actuators (the long, lateral, and medial heads of the triceps brachii; the long and short heads of the biceps brachii (LHB, SHB); and the brachioradialis (BRD)) as well as a CAD model of the exoskeleton robot with a joint motor. In real-world laboratory settings, researchers may use other types of rehabilitation robots to apply controlled rotational motion to the elbow joint \cite{schmitReflexTorqueResponse1999, kooNeuromusculoskeletalModelSimulate2006, guo2022practical}. Nonetheless, the effect of other robots would be similar as the one of the exoskeleton in the simulation.

The original simulation tool \cite{yu2024WCCI} incorporates two controllers: a robot PID controller and a human RL controller, which respectively modulate the motor of the exoskeleton and the muscle activation of the elbow. The elbow musculoskeletal model, guided by its RL controller, can simulate movements similar to those of a healthy real elbow. Meanwhile, the robot PID controller enables the exoskeleton to precisely track trajectories in position control or apply specific external torques in force control. This parallel control of the elbow and robot facilitates the simulation of their physical interactions in typical rehabilitation experiments, such as perturbation-based joint impedance identification experiments \cite{yu2024WCCI}.

Building on this framework, we developed and integrated a stretch reflex controller into the simulation control loop. During each control cycle, the stretch reflex controller receives inputs on muscle fiber length, velocity, and force, simulating proprioceptor functions within muscles. It then generates a muscle reflex response based on proprioceptive feedback, simulating the action of spinal motor neurons. This reflex is calculated using an equation-based model that describes the mechanics of muscle spasticity. The resulting spasticity-related muscle reflexes are then fused with the muscle activation outputs from the human RL controller, producing the final control commands for muscle actuators in the simulation environment. The muscle actuators produce torques around the elbow joint following the Hill-type musculotendon dynamic model \cite{MyoSuite2022}. When the human RL controller is disabled to simulate a completely relaxed elbow with no voluntary control, the muscle activations are driven solely by the stretch reflex controller, allowing for an isolated analysis of reflex-driven muscle behaviour.

\subsection{Spasticity model and stretch reflex controller}
Although no universal model can precisely describe the connections between spasticity pathophysiology and clinical symptoms, it is generally accepted that the abnormal stretch reflex in spasticity can be modelled as a dynamic function of exaggerated muscle excitation driven by proprioceptive feedback \cite{chaQuantitativeModelingSpasticity2020}. Spasticity is thought to arise from reduced inhibition of stretch reflex within the spinal cord, along with an increased reflex gain due to changes in the intrinsic properties of motoneurons \cite{burkePathophysiologySpasticityStroke2013}. On the one hand, the neural excitation underlying the stretch reflex originates from the firing rates of muscle spindles, which are influenced by both muscle length and muscle velocity feedback. For a given spindle firing rate, a reduction in inhibitory inputs to the motoneurons can lead to higher levels of excitation \cite{trompettoPathophysiologySpasticityImplications2014}. This heightened excitation causes the stretch reflex to be elicited at lower velocities and shorter stretch distances. On the other hand, following a neurological injury, disrupted inputs to the motoneuron pool would activate persistent inward currents, which amplify inputs to motoneurons and cause self-sustained firing \cite{burkePathophysiologySpasticityStroke2013}. This process results in an exaggerated motoneuron response to incoming signals, contributing to the excessive reflex activity characteristic of spasticity. 

In addition, there are evidences showing that the spindle firing is proportional to muscle force and force derivative during passive stretches \cite{blumForceEncodingMuscle2017}, so some researchers believed that spasticity could be related to the exaggerated response to spindle firing turned by the feedback of muscle force and its first-time derivative \cite{falisseSpasticityModelBased2018}. Given that, we implemented and tested force-, velocity- and length-dependent models of muscle spasticity and their combinations to comprehensively compare their capability to describe the characteristics of passive stretch resistance observed in elbows with spasticity.

To describe spasticity models incorporating different feedback mechanisms, we adapted the hybrid model equation for the muscle activation of a motoneuron proposed in \cite{veerkamp2023predicting}:
\begin{equation} \label{Eq:activation}
        E(t) = C + G_l \cdot R_l(t) + G_v \cdot R_v(t) + G_f \cdot R_f(t)
\end{equation} 
where $E(t) \leq 1$; $C$ represents the supraspinal drive modulated by the human RL controller, which is set to zero when simulating a completely relaxed elbow; $R_l$, $R_v$, and $R_f$ denote the reflex activations contributed by muscle length, velocity, and force feedbacks within the afferent pathway; $G_l$, $G_v$, and $G_f$ are the reflex gains of spasticity models, which can disable the corresponding model when set to 0. Similar to the modelling method of reflex thresholds used by He et al. \cite{heStretchReflexSensitivity1998, fengNeuromuscularModelStretch1998}, the feedback-dependent reflex activations are calculated through normalisation functions with thresholds of corresponding feedbacks:
\begin{align} \label{Eq:length}
    \! R_l(t+\tau) & =     
        \begin{cases}
            (l(t)-l_t)/(l_0 + l_r), & \text{if } l(t) \geq l_t \\
            0, & \text{if } l(t) < l_t 
        \end{cases}\\
    l_t & = l_0 + \lambda_l \cdot l_r , \text{ } 0 \leq \lambda_l \leq 1 
\end{align}
\begin{align} \label{Eq:velocity}
    \! R_v(t+\tau) & = 
        \begin{cases}
            A \cdot (v(t) - v_t)/v_{max}, & \text{if } v(t) \geq v_t \\
            0, & \text{if } v(t) < v_t
        \end{cases}\\
    v_t & = \lambda_v \cdot v_{max}, \text{ } 0 \leq \lambda_v \leq 1
\end{align}
\begin{align} \label{Eq:force}
    \! R_f(t+\tau) & = 
    \begin{cases}
        A \cdot (f(t) - f_t)/f_{max}, & \text{if } f(t) \geq f_t \\
        0, & \text{if } f(t) < f_t
    \end{cases}\\
    f_t & = \lambda_f \cdot f_{max}, \text{ } 0 \leq \lambda_f \leq 1
\end{align} 
\begin{align}
    A & = 3\cdot l_r/(l_0 + l_r)
\end{align}
where $l(t)$, $v(t)$, and $f(t)$ are the proprioceptive feedbacks of fibre length, velocity, and tension force; $l_t$, $v_t$, and $f_t$ are the reflex thresholds of muscle stretch length, velocity, and force; $\lambda_l$, $\lambda_v$, and $\lambda_f$ are the factors to adjust the thresholds of the stretch reflex; $l_0$ is the starting muscle length of stretches; $l_r$ is the range of muscle stretch; $v_{max}$ is the maximal contracting velocity of a muscle, which was set as 1.5 times the optimal resting length of the muscle per second \cite{XMLReferenceMuJoCo}; $f_{max}$ is the maximal muscle force, which was set as 1.2 times the static maximal force \cite{XMLReferenceMuJoCo}; $\tau$ is the stretch reflex delay, which was set as 30 $ms$ \cite{kooNeuromusculoskeletalModelSimulate2006}; $A$ is the scaling factor that balances the reflex magnitudes of different models at a same level. In our simulation tool, we deployed three separate stretch reflex controllers based on the aforementioned feedback-dependent spasticity model to the LHB, SHB, and BRD muscles, respectively.

\subsection{Robot-assisted elbow stretch experiment}
The constant-velocity elbow stretch experiment is a typical biomechanical measurement method used in spasticity assessment research \cite{guo2022practical, mcphersonBiomechanicalParametersElbow2019, kooNeuromusculoskeletalModelSimulate2006, schmitReflexTorqueResponse1999}. This method involves using a robotic device to exert slow to fast constant-velocity stretching on the relaxed limb joint while recording the joint position and resistance over time. A fundamental assumption of this method is that when the stretching speed exceeds a threshold value, a spastic elbow will exhibit velocity-dependent stretch reflexes, demonstrating a lower reflex threshold and greater resistance compared to healthy individuals \cite{mcphersonBiomechanicalParametersElbow2019, kamperEffectMuscleBiomechanics2001}. In the early study proposing this method, researchers used a motor-driven beam connected to the forearm to apply ramp-and-hold extension and flexion movements to the elbow joint \cite{schmitReflexTorqueResponse1999}. A ramp-and-hold movement includes a start phase that fixes the elbow at the initial position, a ramp phase that moves the elbow to the target position at a constant velocity, and a hold phase that stops the elbow at the target position. In recent studies by Guo et al. \cite{guo2022practical}, this method was followed, but they used a planar forearm rehabilitation robot to conduct the experiment.

\begin{figure}[t]
\centering
\includegraphics[scale=0.35]{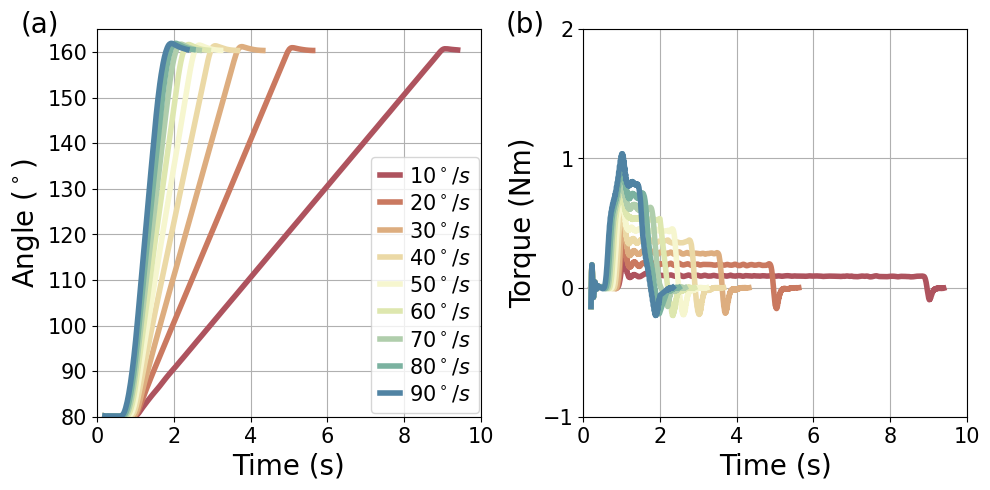}
\caption{Simulation data of the robot-assisted constant-velocity elbow stretch experiments: a) The angle profiles of the ramp-and-hold extension movements; b) The driving torques collected in the simulated elbow stretch experiment.}
\label{fig:raw}
\vspace{-5mm}
\end{figure}

We reproduced Guo's experimental protocol in the simulation environment. The arm was attached to the robot with an initial posture of approximately 90$^\circ$ shoulder elevation and 70$\sim$90$^\circ$ elbow angle (completely extending elbow is 180$^\circ$). The human RL controller was disabled to simulate a relaxed elbow while the robot extended the elbow from the initial position to full extension at nine different angular velocities (ranging from 10 to 90$^\circ/s$). A sample of the experimental data is depicted in Fig.\ref{fig:raw}. The stretch experiment was repeated on the virtual elbows with different model parameters in the simulation tool. The model parameters of the stretch reflex controllers set for the simulation are provided in TABLE \ref{table:spasticity}. Through setting different combinations of gains and thresholds, the general spasticity model described by Eq.\ref{Eq:activation} was transformed into the length-dependent model, velocity-dependent model, force-dependent model, and hybrid model of length and velocity muscle feedbacks. The muscle parameters used in all simulation experiments are the same, as listed in TABLE \ref{table:muscle}, which were calculated by the internal parameters of the MyoSuite elbow model.

\begin{table}[t]
    \centering
    \caption{Spasticity parameters set in the simulation}
    \begin{tabular}{|c|c|c|c|c|c|c|}
    \hline
    Model & $G_l$ & $G_v$ & $G_t$ &$\lambda_l$& $\lambda_v$&$\lambda_f$\\
    \hline \hline
    Len. & $1\sim3$ & 0 & 0 & 0.1 & 0 & 0 \\
    \cline{1-7}
    Vel. & 0 & $1\sim3$ & 0 & 0 & 0.1 & 0 \\
    \cline{1-7}
    For. & 0 & 0 & $1\sim3$ & 0 & 0 & 0.1 \\
    \cline{1-7}
    Hybrid & $1\sim2$ & $1\sim2$ & 0 & $0.1\sim0.4$ & $0.1\sim0.4$ & 0 \\
    \hline
    \end{tabular}
    \label{table:spasticity}
\end{table}

\begin{table}[t]
    \centering
    \caption{Muscle parameters set in the simulation}
    \begin{tabular}{|c|c|c|c|c|}
    \hline
    Muscle & $l_0 [m]$ & $l_r [m]$ & $v_{max} [m/s]$ & $f_{max} [N]$ \\
    \hline \hline
    LHB & 0.36 & 0.054 & 0.18 & 729 \\
    \cline{1-5}
    SHB & 0.28 & 0.054 & 0.20 & 508 \\
    \cline{1-5}
    BRD & 0.12 & 0.024 & 0.13 & 1171 \\
    \hline
    \end{tabular}
    \label{table:muscle}
\end{table}

\begin{figure*}[h]
\centering
\includegraphics[scale = 0.3]{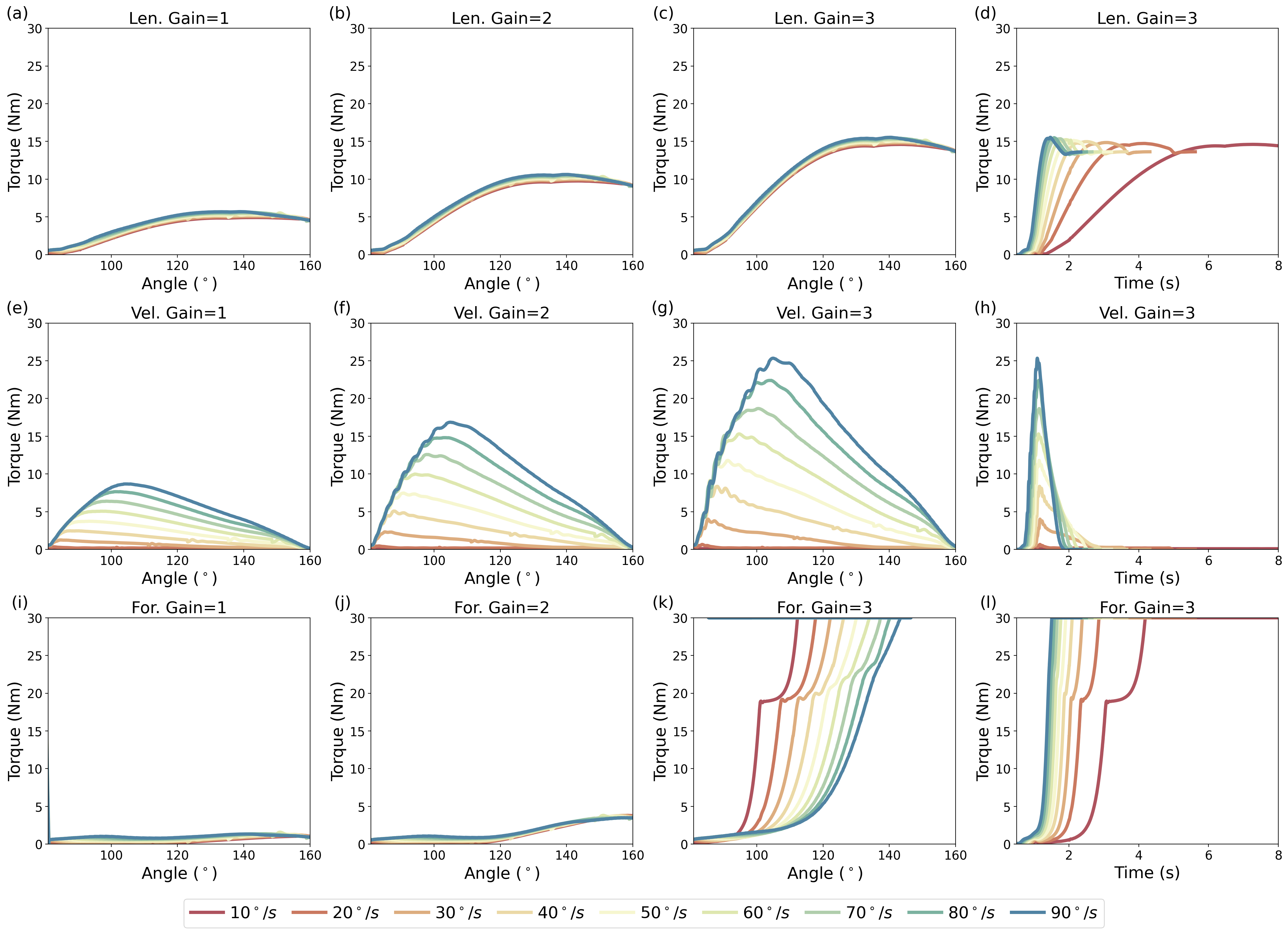}
\caption{(a-c) Reflex torque curves plotted against joint angle for the length-dependent spasticity model, with reflex gains ranging from 1 to 3. (e-g) Reflex torque curves plotted against joint angle for the velocity-dependent spasticity model, with reflex gains ranging from 1 to 3. (i-k) Reflex torque curves plotted against joint angle for the force-dependent spasticity model, with reflex gains ranging from 1 to 3. (d, h, l) Evolution of reflex torques over time for the length-, velocity-, and force-dependent spasticity models, with a reflex gain of 3.}
\label{fig:results}
\vspace{-5mm}
\end{figure*}

In the simulated experiments, virtual sensors within the simulation environment recorded the elbow joint angles $\theta$ and the driving torques $T_{total}$. According to Newton's third law, this driving torque represents the stretch resistance of the joint, which can be described as a function of joint angle. For a spastic elbow joint, the stretch resistance is divided into passive resistance $T_p$ contributed by the mechanical property of body inertia and soft tissues (muscles, tendons, ligaments, and fascia, etc.) and reflex resistance $T_s$ elicited by spasticity models \cite{schmitReflexTorqueResponse1999}. The passive resistance can be further decomposed into the elastic torque $T_e$ related to the joint displacement, viscous torque $T_v$ dependent on the joint velocity, and inertial torque $T_i$ \cite{kooNeuromusculoskeletalModelSimulate2006}, which accounts for accelerating and decelerating the body inertia:
\begin{align} \label{torque}
    \! T_{total}(\theta) & = T_p(\theta) + T_s(\theta) \\
    \! T_p(\theta) & = T_e(\theta) + T_v(\theta) + T_i(\theta)
\end{align} 
During the constant-velocity phase of each passive extension movement, the acceleration is zero, so inertial resistance is zero. The reflex torque elicited by the elbow flexors over this constant angular velocity range can be estimated by subtracting the passive elastic and viscous torques from the driving torque \cite{kooNeuromusculoskeletalModelSimulate2006}. The viscous torque is relatively small compared to the reflex-induced torque (see Fig.\ref{fig:raw}(b) and Fig.\ref{fig:results}), which does not influence our analysis results. The passive elastic torque was estimated by the driving torque recorded at the lowest extension velocities when disabling the reflex controller. We analysed the performance of different spasticity models by examining the reflex torque curves obtained in the simulation experiments.


\section{Results}
We compared the reflex torque curves of the length-, velocity-, and force-dependent spasticity models, with reflex gains incrementally varying from 1 to 3. The outputs of the length-dependent spasticity model are shown in Fig.\ref{fig:results} (a-d). As the elbow muscles are lengthened, the reflex torques gradually increase, reaching a plateau or slightly declining. This behaviour aligns with the intuitive understanding of a length-dependent model: as muscle length initially increases during stretching and then stabilises, the muscle excitation—proportional to fibre length—follows a similar pattern. The slight reduction at the end is attributable to the decreasing muscle velocity, as the force in a Hill-type muscle model is determined by the product of the length-tension and velocity-tension functions.

Fig.\ref{fig:results} (e-h) presents the simulated results of the stretching experiment using the velocity-dependent spasticity model. It is evident that the magnitudes of reflex torques depend on velocity, with the joint angle of peak torque increasing as velocity increases. This may be due to the greater acceleration distance involved in faster stretching under a relatively consistent acceleration time. Additionally, joint resistance decreases after reaching the peak, even though the joint continues to move at a constant velocity. This reduction may be attributed to the decreasing stretch velocity of muscles as their moment arms change.

\begin{figure}[t]
\centering
\includegraphics[scale=0.35]{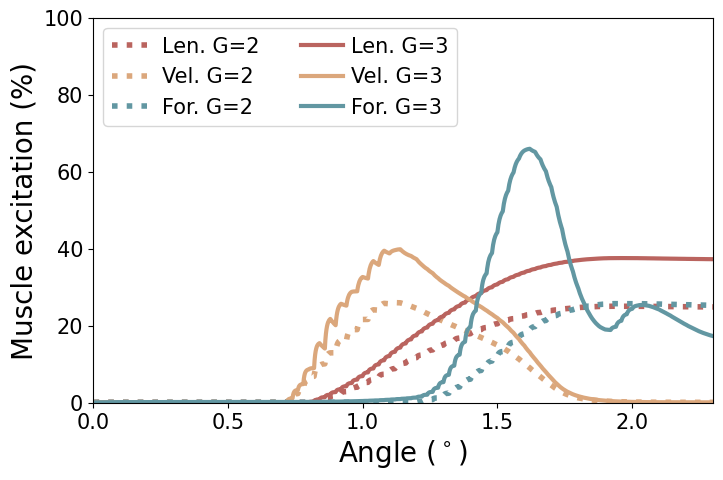}
\caption{Muscle excitations of the long head of the biceps brachii modulated by the length-, velocity-, and force-dependent spasticity models under a 90$^\circ/s$ elbow stretch.}
\label{fig:excitation}
\vspace{-5mm}
\end{figure}

Fig.\ref{fig:results} (i-l) shows the results of the extension experiment for the force-dependent spasticity model. When the gain is small, the force-dependent model induces an increase in resistance during the second half of stretching. However, when the spasticity gain is large, the reflex torque rapidly increases towards the limit value of the muscle force. We further plotted Fig.\ref{fig:excitation} to compare the muscle excitation levels regulated by different models when $G$=2 and $G$=3. It can be observed that at $G$=2, the excitation peak values generated by the three models are quite close. However, at $G$=3, the force-dependent model shows a sharp increase and fluctuation in muscle excitation. It may be explained by the mathematical description of the model. The spasticity model causes greater reflex muscle forces, and greater muscle forces also produce more violent reflexes, so the elbow system cannot reach a stable status.

\begin{figure}[t]
\centering
\includegraphics[scale=0.35]{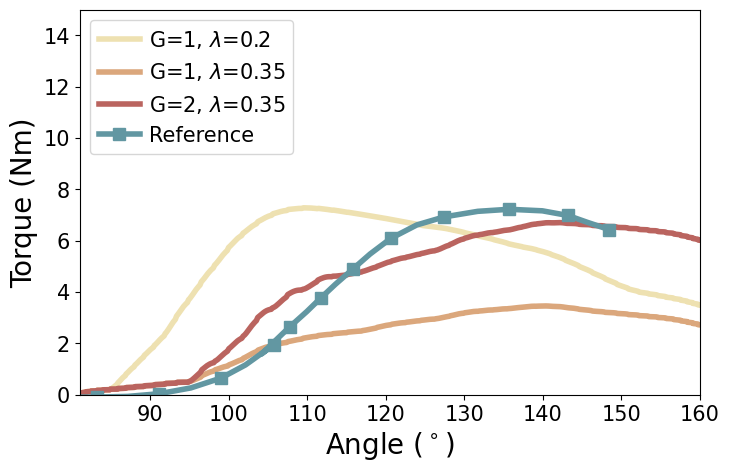}
\caption{Reflex torque curves plotted against joint angle for the hybrid spasticity model with different model parameters ($G$: gains, $\lambda$: threshold factor) at 90$^\circ/s$. The reference torque curves are extracted from McPherson et al. \cite{mcphersonBiomechanicalParametersElbow2019}.}
\label{fig:fitting}
\vspace{-5mm}
\end{figure}

Previous studies have clearly analysed the trend of stretch reflex resistance in spastic elbows, where the most common pattern of resistance profiles is an initial increase followed by a plateau phase \cite{schmitReflexTorqueResponse1999, kamperEffectMuscleBiomechanics2001, mcphersonBiomechanicalParametersElbow2019}. The rate of increase in resistance relative to joint angle (commonly referred to as reflex stiffness) and the onset of abnormally elevated resistance (typically called the reflex angle or catch angle) are both related to the severity of spasticity and the stretching velocity. In our simulation results, the length-dependent model does not exhibit velocity dependence, while the velocity-dependent model lacks a plateau phase in the resistance curve. As a result, we ultimately adopted a length-velocity hybrid spasticity model and conducted simulation experiments using different model parameters listed in TABLE \ref{table:spasticity}. The gains and thresholds for length and velocity were set to identical values ($G_l$=$G_v$, $\lambda_l$=$\lambda_v$).  As shown in Fig.\ref{fig:fitting}, When the gains are kept constant and the thresholds are increased, the onset of torque increase occurs later, and the magnitude of the torque peak decreases. Conversely, increasing the gains while keeping the thresholds unchanged results in an obvious increase in the peak magnitude of the reflex torque curve. With appropriately selected model parameters ($G$=$2$, $\lambda$=$0.35$), the overall trend of joint resistance closely aligns with experimental data reported in the literature \cite{mcphersonBiomechanicalParametersElbow2019}.


\section{Discussion}
In this study, we proposed a simulation framework for spasticity modelling through passive stretch experiments based on the simulation of physical interaction between a rehabilitation robot and a human elbow. Unlike previous simulation of spasticity models, the realisation of the framework does not rely on motion or electromyography data from real-life experiments, nor does it generate the simulation depending on optimal fitting to limited sample data. Instead, it leverages the advantages of the robotic simulator, combined with validated spasticity models, to perform forward simulations purely according to the physiological mechanism. 

Our simulation examined length-, velocity-, and force-dependent models separately, all of which demonstrated unique characteristics, but none of the models alone could capture the full passive resistance curve observed in elbow stretches of spastic patients. It was only when combining both velocity and length dependencies that the hybrid model displayed passive resistance patterns seen in a real spastic patient' elbow. This can be explained by the physiological origin of the stretch reflex, the muscle spindle, which contains Ia and II sensory fibres sensitive to muscle velocity and length, respectively. The activation of these sensory fibres in response to external stretch increases the afferent firing rate to the motoneuron, which may elicit the stretch reflex \cite{kooNeuromusculoskeletalModelSimulate2006}. Previous experimental evidence has demonstrated that the muscle spindle in a spastic patient is not more sensitive than in a normal individual, and spasticity is more closely associated with a reduction in the inhibition of motoneuron excitability within the spinal cord \cite{burkePathophysiologySpasticityStroke2013, trompettoPathophysiologySpasticityImplications2014}. Therefore, we infer that the weakening of afferent-related neural inhibition may lead to the abnormal stretch reflex exhibiting length- and velocity-dependent excitation originating from the muscle spindle.

In Falisse et al.'s study \cite{falisseSpasticityModelBased2018}, spasticity models were described by state equations that incorporated reflex threshold and gain, with state variables of muscle feedbacks (length, velocity, acceleration, force, and force derivative), while the time course of muscle excitation was the simulation output. Optimisation of model parameters was conducted to make the simulated time course of muscle excitation fit the electromyography data from real patients. However, the muscle force input of the model in their optimisation framework was generated from the real electromyography data, which might result in a fitting bias that the force-dependent model has better optimisation results than the model based on other feedback. In contrast, our study directly simulates the force-dependent model without employing any optimisation techniques, allowing us to observe the model's behaviour without any pre-set bias from fitted data. Our findings reveal that models based solely on force variables lead to muscle system instability, producing physiologically unrealistic results.

Veerkamp et al. compared the velocity-dependent and force-dependent spasticity models in reproducing spastic patients' gait patterns \cite{veerkamp2023predicting}. Similar to our findings, the velocity-dependent model outperforms the force model under their simulation goals. Their simulation method consists of two layers of internal and external optimisation loops, and therefore the simulation is computationally intensive, so they failed to validate the spasticity model with multiple feedback combinations. In terms of computational cost, our proposed simulation tool is advantageous. It is true that adding optimisation of model parameters to our simulation tool to make it fit to the experimental data will introduce new computational costs. However, our simulation framework only requires one optimisation loop. Moreover, the simulation efficiency of MyoSuite is higher than OpenSim \cite{MyoSuite2022}. Therefore, we believe that the present simulation tool has the potential to improve the computational efficiency of spasticity model simulation.

\section{Conclusion}
It is undeniable that, due to the lack of clinical trial data, this study does not propose a rigorously validated spasticity model. The musculoskeletal model of the elbow requires improvement as it excludes the Brachialis muscle. However, our simulation framework provides new approaches and insights into some of the ongoing debates in spasticity modelling. The framework enables easier comparison of stretch resistance at the elbow joint across different types of spasticity models. The preliminary qualitative analysis results demonstrate the superiority of the length- and velocity-dependent hybrid spasticity model in explaining the abnormal elbow resistance caused by spasticity. Furthermore, while not covered in this paper, the effects of spasticity on active human motion can also be simulated, as the simulation tool includes a controller for active motion. In future research, one focus will be on gathering some real-patient data to develop a more precise and reliably validated spasticity model, thereby enhancing the simulation fidelity for spastic patients. 



\bibliographystyle{IEEEtran}
\bibliography{ref_robot}

\end{document}